\documentclass[10pt,twocolumn,letterpaper]{article}

\usepackage{cvpr}              % To produce the CAMERA-READY version
\usepackage{url}
\usepackage{paralist}

\definecolor{cvprblue}{rgb}{0.21,0.49,0.74}
\usepackage[pagebackref,breaklinks,colorlinks,allcolors=cvprblue]{hyperref}

\def\paperID{2928} % *** Enter the Paper ID here
\def\confName{CVPR}
\def\confYear{2026}

\title{Multi-view Crowd Tracking Transformer with View-Ground Interactions \\ Under Large Real-World Scenes}

\author{
\begin{tabular}{c}
Qi Zhang$^{1}$ \quad
Jixuan Chen$^{1}$ \quad
Kaiyi Zhang$^{1}$ \quad
Xinquan Yu$^{1}$ \\
Antoni B. Chan$^{2}$ \quad
Hui Huang$^{1}$\thanks{Corresponding author} \\
$^{1}$College of Computer Science and Software Engineering, Shenzhen University, China \\
$^{2}$Department of Computer Science, City University of Hong Kong, China \\
{\tt\small \{qi.zhang.opt, chen.jixuanstu, zhangky1999, xinquanyu2619\}@gmail.com,}\\
{\tt\small abchan@cityu.edu.hk, hhzhiyan@gmail.com}
\end{tabular}
}

\begin{document}                                                                   
\maketitle
\begin{abstract}
Multi-view crowd tracking estimates each person's tracking trajectories on the ground of the scene. Recent research works mainly rely on CNNs-based multi-view crowd tracking architectures, and most of them are evaluated and compared on relatively small datasets, such as Wildtrack and MultiviewX. Since these two datasets are collected in small scenes and only contain tens of frames in the evaluation stage, it is difficult for the current methods to be applied to real-world applications where scene size and occlusion are more complicated. In this paper, we propose a Transformer-based multi-view crowd tracking model, \textit{MVTrackTrans}, which adopts interactions between camera views and the ground plane for enhanced multi-view tracking performance. Besides, for better evaluation, we collect and label two large real-world multi-view tracking datasets, MVCrowdTrack and CityTrack, which contain a much larger scene size over a longer time period. Compared with existing methods on the two large and new datasets, the proposed MVTrackTrans model achieves better performance, demonstrating the advantages of the model design in dealing with large scenes. We believe the proposed datasets and model will push the frontiers of the task to more practical scenarios, and the datasets and code are available at: \url{https://github.com/zqyq/MVTrackTrans}.
\end{abstract}    
\section{Introduction}
\label{sec:intro}
%\kynote{context consistency of model name and dataset attributes have already been checked}
%\zqnote{Draw a figure showing the general difference of the model and datasets here.}

% \begin{figure*}
%   \centering
%   \begin{subfigure}{0.68\linewidth}
%     \fbox{\rule{0pt}{2in} \rule{.9\linewidth}{0pt}}
%     \caption{An example of a subfigure.}
%     \label{fig:short-a}
%   \end{subfigure}
%   \hfill
%   \begin{subfigure}{0.28\linewidth}
%     \fbox{\rule{0pt}{2in} \rule{.9\linewidth}{0pt}}
%     \caption{Another example of a subfigure.}
%     \label{fig:short-b}
%   \end{subfigure}
%   \caption{Example of a short caption, which should be centered.}
%   \label{fig:short}
% \end{figure*}

\begin{figure}[htbp]
    \centering
    \includegraphics[width=\linewidth]{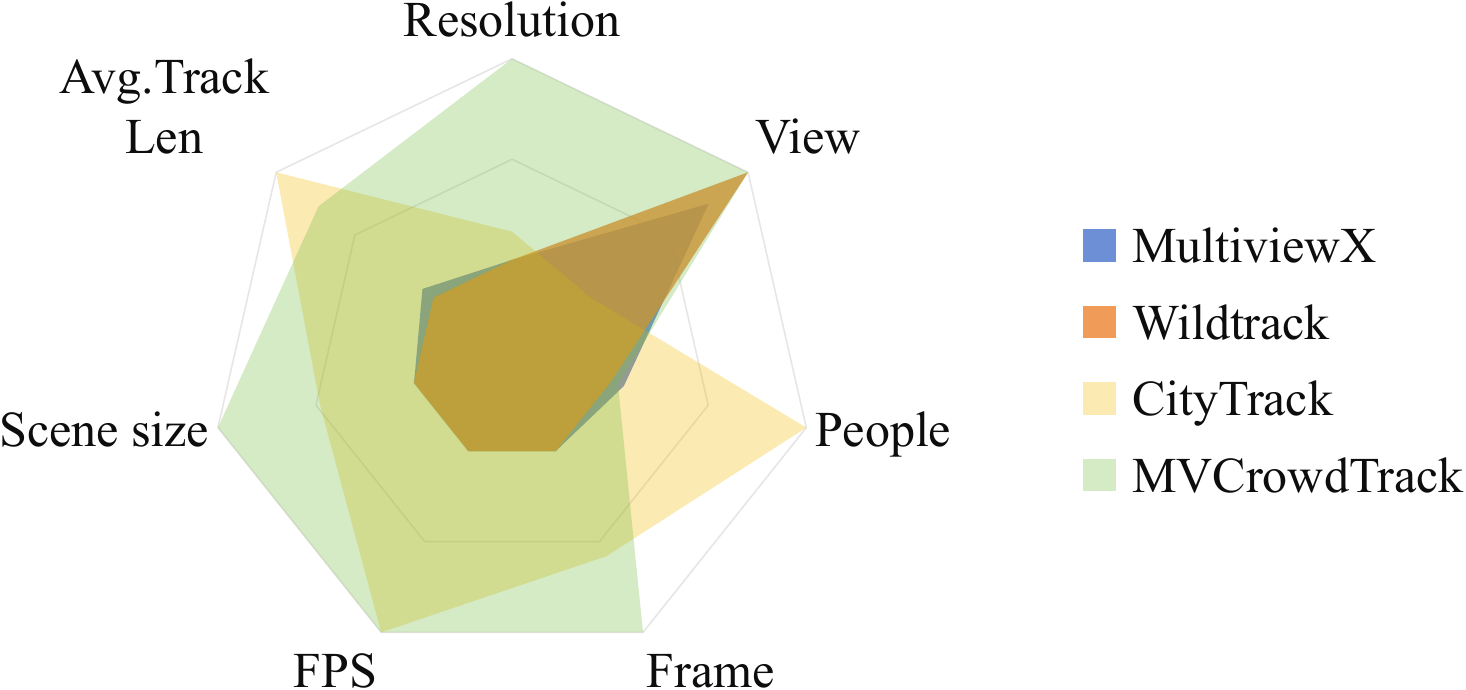}
    \vspace{-0.3cm}
    \caption{Comparison of the proposed MVCrowdTrack and CityTrack datasets with existing multi-view tracking datasets MultiviewX and Wildtrack. Our proposed datasets are superior in various aspects, featuring larger scene coverage, longer average track lengths, and a greater number of people, providing a more comprehensive benchmark for multi-view crowd tracking.}
    \label{fig:comparison}
    \vspace{-0.6cm}
\end{figure}

% \begin{figure*}[htbp]
%     \centering
%     \includegraphics[width=\textwidth]{sec/figures/canghai_vis_cropped.pdf}
%     \caption{Examples from the Canghai dataset. The figure shows camera frames and the corresponding camera layout.}
%     \label{fig:canghai}
% \end{figure*}

Multi-view crowd tracking estimates each person's tracking trajectories for a time period on the ground of the scene by accumulating the information from multiple synchronized and calibrated cameras. It can be applied to many applications, such as crowd management \cite{10.1007/978-3-031-72943-0_2}, public transportation \cite{zhao2021mdlf}, autonomous driving \cite{li2024bevformer}, \textit{etc}.

Most of the recent works \cite{Teepe_2024_CVPR, Teepe_2024_WACV} evaluate their methods and compare with others on relatively small datasets, such as Wildtrack \cite{chavdarova2018wildtrack} and MultiviewX \cite{hou2020multiview}, which are recorded on small scenes and only consist of hundreds of frames in total. Thus, the existing works may not be well applied to real-world scenarios, since real-world multi-view crowd tracking may happen on large scenes with a large crowd and severe occlusions, and over a long time period. Therefore, to study the multi-view crowd tracking task under more difficult real scenes is in demand in the area.

%\zqnote{Draw a figure showing the detailed model architecture.}
\begin{figure*}[htbp]
    \centering
    \includegraphics[width=0.9\linewidth]{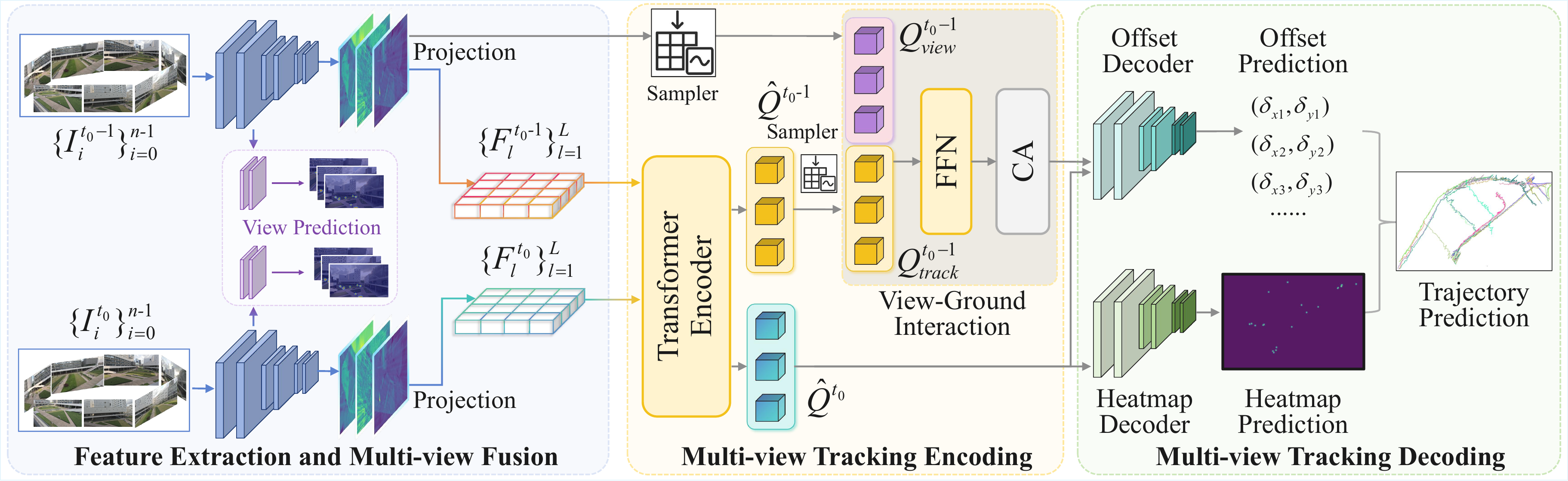}
    \vspace{-0.2cm}
    \caption{The overall pipeline of MVTrackTrans consists of Feature Extraction and Multi-view Fusion, Multi-view Tracking Encoding, and Multi-view Tracking Decoding. Consecutive multi-view frames are processed by a shared feature extractor and projected onto the ground plane to obtain fused ground features. These features are then encoded by a Transformer Encoder to generate track queries from the previous timestep and current frame queries from the current one. A View-Ground Interaction module is proposed to refine track queries by combining feature queries from both camera views and the ground plane. Finally, the Offset Decoder predicts temporal offsets, while the ground Heatmap Decoder outputs the current location detection results, which are combined together to obtain the predicted trajectories.}
% The consecutive multi-view frames are fed into a shared feature extractor to obtain camera-view features, which are then projected onto the ground plane to generate fused ground-plane representations. These ground features are passed through a Transformer Encoder to enhance spatial representations and produce track queries from the previous timestep along with detection queries from the current one. The track queries are discretized by sampling specific track locations, while the View-Ground Interaction module further samples track locations from camera views and concatenates them with ground-plane samples, yielding more representative track queries. Finally, the track and detection queries are fed into the Offset Decoder to estimate temporal offsets between frames, and the detection queries are passed to the Detection Decoder to predict the current timestep’s detection results. 
    \label{fig:pipeline}
    \vspace{-0.5cm}
\end{figure*}

In this paper, we first address the dataset and evaluation issues by collecting and labeling two large real-world multi-view crowd tracking datasets: MVCrowdTrack and CityTrack. MVCrowdTrack is a newly collected dataset captured in a campus environment, 
containing 4,122 frames from seven synchronized camera views.  
CityTrack is constructed based on the existing CityStreet dataset \cite{zhang2019wide}, 
which originally contains 500 multi-view frames sampled at 2\,fps.  
We re-sampled the videos at 4\,fps and re-annotated them for multi-view tracking, 
resulting in 2,588 frames with denser temporal continuity. 
As shown in Figure \ref{fig:comparison}, compared to datasets used in the state-of-the-art methods \cite{Teepe_2024_WACV, Teepe_2024_CVPR}, \textit{e.g}., Wildtrack and MultiviewX, the proposed datasets contain larger scenes, more frames with longer video duration, and more number of trajectories (CityTrack) with longer average trajectory length.
Overall, the proposed datasets are more suitable for studying the multi-view crowd tracking task under difficult real-world scenes.
%since they are collected on real-world scenes with large scene sizes and occlusions.

Besides, it is noticed that most SOTAs adopt CNNs-based architecture for the multi-view crowd tracking task, and seldom explore more advanced architectures such as transformers. In contrast, we propose a Transformer-based multi-view crowd tracking model, denoted MVTrackTrans, which performs the multi-view crowd tracking on the ground with transformers. Furthermore, MVTrackTrans adopts interactions between camera views and the ground plane for better multi-view tracking. As illustrated in Figure~\ref{fig:pipeline}, the whole pipeline of the proposed MVTrackTrans model consists of three steps:
(1) \textbf{Feature extraction and Multi-view Fusion}: Each camera view's features of the two neighboring frames are extracted with a shared feature extractor and then projected to the ground plane for multi-view fusion to obtain the ground representation.
(2) \textbf{Multi-view Tracking Encoding}: To capture temporal dynamics, the ground representations from the previous and current frames are encoded by a deformable encoder. From the previous-frame ground representation, a set of discrete track queries is sampled to represent the tracked entities in the ground space. A \textit{view-ground interaction} module is proposed, where these queries interact with the previous-frame multi-view detection features via cross-attention, enabling cross-view and temporal information exchange.
(3)\textbf{ Multi-view Tracking Decoding}: The refined queries are then fused with the current frame's ground representation to jointly decode motion offsets and detection heatmaps of the crowd, from which the final tracking results are obtained.

% the $S$ RGB images ($S \times 3 \times H_i \times W_i$) are first processed by a ResNet backbone to extract multi-level features from the first three stages. These single-view features are projected into the ground space via calibrated camera parameters, and multi-scale ground features from all views are aggregated through convolutional fusion to form unified ground representations $\{F^{t_0}_{\mathrm{scene},l}\}_{l=1}^{3}$. To capture temporal dynamics, both the ground features and multi-view features from the previous frame are encoded by a deformable encoder, from which discrete \textit{track queries} are sampled to represent tracked entities in the ground space. These queries interact with the previous-frame multi-view detection features via self-attention, enabling cross-view and temporal information exchange. The refined queries are then fused with the current-frame ground features to jointly decode motion offsets and detection heatmaps, from which the final tracking results are obtained.

As far as we know, this is the first time that multi-view crowd tracking is put forward to much larger real-world scenes with more crowds and longer time periods, and a transformer-based multi-view crowd tracking model is presented. We believe the proposed datasets and MVTrackFormer model will push the frontiers of the area to more practical scenarios. Our main contributions are as follows.
\begin{compactitem}
  \item We propose two large real-world datasets, MVCrowdTrack and CityTrack, for multi-view crowd tracking. Compared to existing datasets widely used in the area, the proposed two datasets are much more difficult and more suitable for real-world research, which shall advance the area to more practical applications.

  \item We propose a transformer-based multi-view crowd tracking model, MVTrackTrans, which uses transformer architecture for multi-view crowd tracking instead of CNNs models to deal with the demand for more complicated spatial and temporal association among the crowds in the scene. Besides, MVTrackTrans adopts extra view-ground interactions for better multi-view tracking performance.
  
  \item The experiments validate that the proposed MVTrackTrans model achieves better multi-view crowd tracking performance than comparison methods on the two large real-world datasets, which demonstrates its superiority for the task under complicated scenarios.

\end{compactitem}

\section{Related Work}
\label{sec:relatedwork}
We first review existing multi-view crowd-tracking methods and transformer-based single-view multi-object tracking methods. Then, we also review the related datasets for multi-view crowd-tracking.

%\zqnote{Check arxiv papers, make sure they are cited rightly if published} \kynote{already checked}
\noindent
\textbf{Multi-view Crowd Tracking.}
Multi-view crowd tracking aims to track individuals across multiple cameras. A number of works have been proposed to achieve robust and temporally consistent tracking in multi-view settings. To maintain motion continuity across time, MVFlow \cite{engilberge2023multi} models ground-plane motion flow for occlusion-robust tracking. However, it restricts each person’s movement to a single discrete grid per timestep, which limits its ability to model continuous and realistic human motion dynamics. 
To learn a more discriminative representation for each individual, EarlyBird \cite{Teepe_2024_WACV} builds upon standard BEV-based multi-view detection architectures \cite{zhang2024mahalanobis, hou2020multiview}, integrates an additional ReID module to enhance identity distinction, and leverages a Kalman Filter \cite{kalman1960new} for temporal association across frames.
TrackTacular \cite{Teepe_2024_CVPR} improves on EarlyBird by lifting the BEV representation into 3D space and stacking world representations from adjacent frames to extract temporal information for regressing motion offsets of people. 
Different from previous methods that rely solely on view features or BEV features, DepthTrack \cite{Tran_2025_ICCV} leverages clustered point clouds projected from depth maps to assist BEV-based tracking. By incorporating richer geometric cues such as shape and orientation, the point clouds enhance the model’s ability to distinguish and re-identify individuals more accurately.
Aiming to improve tracking performance in long multi-view videos, MCBLT \cite{Wang_2025_ICCV} employs a Hierarchical Graph Neural Network to perform people association in the BEV space across multiple temporal scales. Each GNN layer focuses on short- and mid-term dependencies, significantly enhancing robustness under long-term occlusions. 
% MITracker \cite{xu2025mitracker} released a new multi-view object tracking benchmark and designed a Spatial-Enhanced Attention module to reduce potential target omissions across different camera views. Despite covering a wider range of object categories than previous datasets, its limited spatial scale and relatively low object density restrict its practicality for real-world multi-view tracking scenarios. 
MVTrajecter \cite{Yamane_2025_ICCV} proposed a framework that utilizes motion and appearance costs across multiple past timestamps, instead of relying solely on the nearest one, to improve trajectory association. By considering information from several previous frames, it achieves more robust tracking performance.

However, most existing approaches rely on CNN-based architectures and rarely explore transformer-based models for multi-view crowd tracking. To address this, MCTR \cite{10972563} proposed a transformer-based framework that iteratively updates global tracking embeddings by interacting with detections from all camera views and introduced a probabilistic association strategy to ensure consistent matching across views and time. \textit{In contrast, while MCTR performs tracking in the original camera views, our MVTrackTrans operates directly in the BEV space and incorporates a View-Ground Interaction module to further enhance the performance}.

\noindent
\textbf{Transformer-based Single-view MOT.} 
% Recent advances in transformer-based multi-object tracking have explored different strategies to improve detection, association, and robustness under challenging scenarios.
% Early works such as GTR \cite{zhou2022global} introduced a novel tracking transformer that employs a global trajectory query to associate detections across all frames. 
Multi-object tracking (MOT) \cite{liu2025sparsetrack, luo2025omnidirectional, chen2025cross, li2025lamot, chamiti2025refergpt, shim2025focusing, gao2025multiple, zhao2025hff} typically uses monocular cameras to follow moving targets. Recent transformer-based approaches have shown substantial gains in detection accuracy, temporal association, and robustness under challenging conditions. For instance, GTR \cite{zhou2022global} proposed a global trajectory query mechanism to associate detections across the entire video sequence.
MeMOT \cite{cai2022memot} developed a large spatio-temporal memory module to store and update object embeddings along trajectories, thereby achieving more robust and consistent tracking performance. 
P3AFormer \cite{zhao2022tracking} proposed detecting and tracking objects at the pixel level, which is particularly beneficial for handling small objects. Similarly, TransCenter \cite{9964258} also leverages dense feature representations for tracking and introduces carefully designed detection and tracking queries to reduce the computational cost of attention over such detailed maps. Modeling object motion is crucial in MOT.
STDFormer \cite{10091152} introduced a purely motion-driven transformer tracking framework that jointly models linear and exponential motion patterns, enabling effective handling of both simple and complex object dynamics. 
To address low tracking performance caused by missed or incorrect detections, BUSCA \cite{Vaquero2024BUSCA} proposed a method to recover lost detections under occlusions by generating potential proposals and formulating the object-proposal association as a multi-choice question-answering task using a decision transformer. 
TGFormer \cite{Zeng_Huang_Pei_2025} addresses occlusions from a different perspective by introducing a group of queries for each object, enabling better capture of appearance variations under varying levels of occlusion.
LA-MOTR \cite{LA-MOTR} proposed a novel learnable association module to iteratively refine the association across multiple attention layers by incorporating relative spatial cues and enabling bidirectional feature interaction between detections and tracklets. 
\textit{Although there are many transformer-based methods for single-view multi-object tracking, transformer architectures for \emph{multi-view} crowd tracking remain largely unexplored. Our MVTrackTrans aims to fill this gap}.

% TransTrack \cite{transtrack} \\
% MOTR \cite{zeng2022motr} Trackformer \cite{meinhardt2022trackformer} \\

\noindent
\textbf{Multi-view Crowd Tracking Datasets.}
% \zqnote{Summarize the existing datasets and compare with ours.}
%\zqnote{Make sure they are consistent with Table 1}
Existing multi-view crowd tracking datasets primarily focus on small- to medium-scale areas with relatively short sequences. 
Table~\ref{tab:dataset_comparison} summarizes the key statistics of representative datasets, including resolution, number of views, people, frames, frame rate, and scene size.
{Wildtrack} \cite{chavdarova2018wildtrack,chen2025cross} is a real-world dataset captured by seven synchronized and calibrated cameras covering an area of $36 \times 12$ m. The camera resolution is $1920 \times 1080$ pixels at 2 fps, and the total sequence length is 400 frames.
{MultiviewX} \cite{hou2020multiview} is a synthetic dataset designed as a virtual replica of Wildtrack. It contains six virtual cameras covering an area of $25 \times 16$ m. The camera resolution, frame rate (2 fps), and sequence length (400 frames) are consistent with Wildtrack. GMVD \cite{vora2023bringing} is also a synthetic multi-view dataset with multiple scenes. Nevertheless, both the scene size and the crowd density are relatively small, remaining largely similar to MultiViewX.
While Wildtrack and MultiviewX are widely adopted in prior multi-view detection \cite{hou2020multiview, hou2021multiview, zhang2024mahalanobis, song2021stacked, qiu20223d, 11094313} and tracking \cite{Teepe_2024_CVPR, Teepe_2024_WACV, Yamane_2025_ICCV} studies, their scene coverage is limited, and the video sequences are relatively short. 

% GMVD \cite{vora2023bringing} is another synthetic multi-view crowd dataset with several distinct scenes. Nevertheless, both the scene size and the crowd density are relatively small, remaining largely similar to those of MultiViewX.
To address these limitations, we introduce two newly collected large-scale real-world multi-view tracking datasets: MVCrowdTrack and CityTrack.
% \textbf{MVCrowdTrack} is collected in a university campus environment using seven synchronized cameras, covering an area of about $120 \times 80$ m, and contains 4,122 frames at 4 fps.
% \textbf{CityTrack} captures a real-world urban intersection scene \cite{zhang2019wide}, covering an area of approximately $64 \times 76$ m. The sequence consists of 2,558 frames recorded at 4 fps, with three cameras.
As shown in Figure \ref{fig:comparison} and Table \ref{tab:dataset_comparison},  MVCrowdTrack and CityTrack are collected in large real-world scenes and provide substantially larger spatial coverage, significantly longer sequences, and larger numbers of crowds than existing benchmarks, enabling more comprehensive evaluation of multi-view tracking methods for real-world applications.

% \kynote{
% MITracker \cite{xu2025mitracker} released a new multi-view object tracking benchmark and designed a Spatial-Enhanced Attention module to reduce potential target omissions across different camera views. Despite covering a wider range of object categories than previous datasets, its limited spatial scale and relatively low object density restrict its practicality for real-world multi-view tracking scenarios. }

\section{Multi-view Crowd Tracking Transformer}

%\zqnote{Talking about the overall architecture of the model and the task settings.}

The multi-view crowd tracking task aims to estimate
all individuals’ trajectories on the ground plane of a scene captured by multiple synchronized and calibrated cameras. In this section, we present our Multi-view Crowd Tracking Transformer framework (as shown in Figure \ref{fig:pipeline}), which unifies ground-plane reasoning and view-level temporal modeling for robust human localization and tracking. The overall architecture consists of three main stages:

(1) \textit{Feature Extraction and Multi-view Fusion}: The multi-view images are first fed into a ResNet backbone to extract multi-level features from the first three stages. The extracted single-view features are then projected into the ground plane using the calibrated camera parameters, producing fused ground features after multi-view fusion. 
%$\{F^{t_0}_{\mathrm{scene},l}\}_{l=1}^{3}$. For each scale $l$, the projected ground features from all views are aggregated via convolutional fusion, resulting in unified multi-scale ground representations $F^{t_0}_{\mathrm{scene},l}$ for subsequent processing.

(2) \textit{Multi-view Tracking Encoding}: To incorporate temporal information, fused ground features from both the previous frame and the current frame are first processed by the deformable encoder to enhance their spatial representation. 
Discrete track queries are then sampled from specific positions within the encoded ground feature map of the previous frame, representing the tracked entities in the ground-plane space. 
Afterward, a View-Ground Interaction module is introduced, where the track queries and the multi-view queries are fused via a cross-attention mechanism.

% A cross-attention mechanism is adopted in the proposed View-Ground Interaction module, 
% where the ground queries serve as queries ($Q$) and the view queries act as keys and values ($K$, $V$).

(3) \textit{Multi-view Tracking Decoding}: Subsequently, the refined track queries and the current-frame ground queries are jointly decoded to estimate the motion offsets of tracked entities in the current frame.
Finally, the current-frame ground queries directly regress the crowd location heatmap. 
By combining the heatmap with the predicted offsets, the final tracking results are obtained. 
See stage details as follows.
%The details of each module are elaborated in the following subsections.

% Overall, the multi-view crowd tracking task aims to estimate
% all individuals’ trajectories on the ground plane of a scene
% captured by multiple synchronized and calibrated cameras.
% The proposed Multi-view Crowd Tracking Transformer framework consists of three main modules:
% (1) feature extraction and multi-view fusion, which extracts image features and projects them into a shared Ground-Plane space;
% (2) view-ground interaction, which integrates temporal and multi-view information for consistent representation learning; and
% (3) transformer encoding and decoding, which estimates human center heatmaps and motion offsets for trajectory prediction.
% The details of each module are described in the following subsections.

\subsection{Feature Extraction and Multi-view Fusion}

We first extract multi-scale single-view features from each camera view using a ResNet backbone~\cite{he2016deep}. 
Let the input multi-view images at timestamp $t_0$ be $\{I_i^{t_0}\}_{i=0}^{n-1}$, 
where $i$ denotes the camera view index and $n$ is the number of cameras. 
The multi-scale single-view features are extracted as: $\{F_{view_i}^{t_0}\}_{l=1}^{L} = \{F(I_i^{t_0})\}_{l=1}^{L}, \quad i = 0,1,\dots,n-1$, 
% \begin{align}
%     F_{i}^{t_0} = F(I_i^{t_0}), \quad i = 0,1,\dots,n-1,
% \end{align}
where $l\!=\!1\!:\!L$ indexes the feature scale and $L$ denotes the number of scales. Each camera then independently passes its multi-scale features through a feature pyramid network (FPN) to produce single-scale view features $\hat{F}_{\mathrm{view}_i}^{t_0}$ for subsequent camera view detection and view feature sampling.

Instead of projecting features to a fixed-height plane, we adopt a multi-height bilinear sampling approach~\cite{harley2023simple}:
\begin{align}
    \begin{pmatrix} u_n, v_n, 1 \end{pmatrix} ^T
    = K [R|T] 
    \begin{pmatrix} x_n, y_n, z_n, 1 \end{pmatrix} ^T.
\end{align}
where each 3D voxel pulls information from the corresponding single-view features. For a voxel with coordinates $(x_n, y_n, z_n)$, its eight vertices are projected to all image planes via the camera projection matrix $K[R|T]$.

The voxel features are sampled from the image features $\hat{F}_{\mathrm{view}_i}^{t_0}$ of all camera views and aggregated across views. 
This ensures that each voxel receives informative features, making the method robust for long-range perception and crowded scenes. 
Finally, the voxel features are collapsed along the height axis and fused across views using convolution, 
resulting in the multi-scale \textit{ground feature} $\{F^{t_0}_{l}\}_{l=1}^{L}$ of the current frame.
In the following stages, the ground features of timestamps $t_0$ and $t_0\!-\!1$ are input into transformer networks for further processing.

%\zqnote{use mathematical language to describe the process of the step, the features of each view, the fused feature representation.}

\begin{figure}[t]
    \centering
    \includegraphics[width=0.6\linewidth]{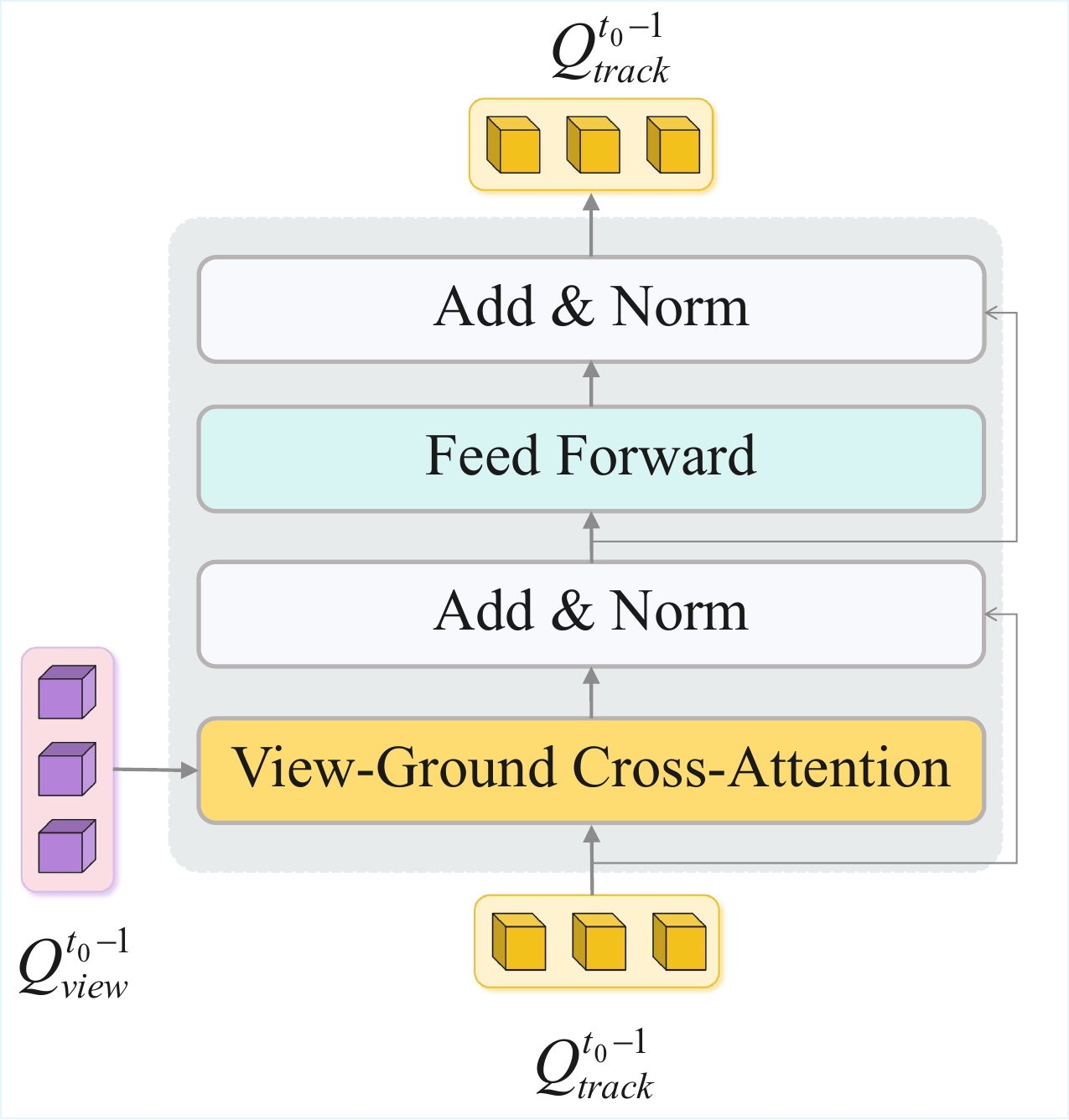}
    \vspace{-0.4cm}
    \caption{The view-ground interaction module details.}
    \label{fig:interation}
    \vspace{-0.6cm}
\end{figure}

\subsection{Multi-view Tracking Encoding}
%\zqnote{The same process, until we got the final output (use a mathematical letter to represent final output and the corresponding GT.)}

We denote the multi-scale ground features at the previous and current timestamps as
$\{F^{t_0-1}_{l}\}_{l=1}^{L}$ and $\{F^{t_0}_{l}\}_{l=1}^{L}$. 
Each set of features is independently processed by the same Transformer Encoder, 
which is implemented using a Multi-scale Deformable Attention Module, 
to aggregate information across multiple scales:
\begin{align}
    \hat{Q}^{t_0\!-\!1} &= \mathrm{Transformer Encoder}(\{F^{t_0-1}_{l}\}_{l=1}^{L}), \\
    \hat{Q}^{t_0} &= \mathrm{Transformer Encoder}(\{F^{t_0}_{l}\}_{l=1}^{L}).
\end{align}
After obtaining the two-frame queries, we perform discrete sampling on the previous-frame query 
$\hat{Q}^{t_0\!-\!1}$ at the past detection locations $(x, y)$ on the ground plane to construct the track queries: $Q^{t_0\!-\!1}_{\mathrm{track}} = \mathrm{SampleQueries}(\hat{Q}^{t_0-1}, (x, y)).$
% \begin{align}
%     Q^{t_0\!-\!1}_{\mathrm{track}} = \mathrm{SampleQueries}(\hat{Q}^{t_0-1}, (x, y)).
% \end{align}

\textbf{View-Ground Interaction.}
As in Figure~\ref{fig:interation}, to better represent each tracked person, we integrate complementary information from both the ground and camera views.
For each camera, we sample a set of view-specific queries from the corresponding view detection features, and then concatenate them across all cameras to form the \textit{view queries}:
\begin{equation}
Q^{t_0\!-\!1}_{\mathrm{view}} = \mathrm{Concat}\big(Q^{t_0\!-\!1}_{\mathrm{view},0}, Q^{t_0\!-\!1}_{\mathrm{view},1}, \dots, Q^{t_0\!-\!1}_{\mathrm{view},n-1}\big),
\end{equation}
where $Q^{t_0\!-\!1}_{\mathrm{view},i}$ denotes the queries sampled from the $i$-th camera view's feature map $\hat{F}_{\mathrm{view}_i}^{t_0}$ in the previous frame.
The track queries ($Q^{t_0\!-\!1}_{\mathrm{track}}$) and view queries ($Q^{t_0\!-\!1}_{\mathrm{view}}$) from the previous frame are first processed by independent feed-forward networks (FFN) to refine their embeddings. 
Then, we adopt a cross-attention mechanism where the track queries serve as the queries $Q$, 
and the view queries sampled from multi-cameras act as the keys $K$ and values $V$: %($K$, $V$):
\begin{align}
    Q^{t_0\!-\!1}_{\mathrm{track}} = 
    \mathrm{CrossAttn}\big(
    \mathrm{FFN}(Q^{t_0\!-\!1}_{\mathrm{track}}),\,
    \mathrm{FFN}(Q^{t_0\!-\!1}_{\mathrm{view}})
    \big).
\end{align}
This design allows each track queries to aggregate visual features from all camera views corresponding to the same tracked individual, 
addressing the limitation that discrete sampling from the previous frame’s ground representation may not fully capture the appearance of the person.

%\zqnote{Add a figure showing the module.}

\subsection{Multi-view Tracking Decoding}

The decoding stage consists of two parallel branches: a heatmap decoder and an offset decoder.

\textbf{Offset Decoder.}  
The offset decoder adopts a standard Multi-scale Deformable Attention (MSDA) structure to model temporal correspondence between consecutive Ground frames. Specifically, given the previous-frame track queries $Q^{t_0-1}_{\mathrm{track}}$, 
the corresponding detection locations $(x^{t_0-1}, y^{t_0-1})$, 
and the current Ground features $\hat{Q}^{t_0}$, the temporal interaction is formulated as:
\begin{align}
    \hat{Q}^{t_0}_{\mathrm{track}} = 
    \mathrm{MSDA}\big(Q^{t_0-1}_{\mathrm{track}},\, \hat{Q}^{t_0},\, (x^{t_0-1}, y^{t_0-1})\big),
\end{align}
where $(x^{t_0-1}, y^{t_0-1})$ serve as reference points to guide the deformable sampling.  
The refined track queries are then passed through a lightweight MLP head 
to predict the motion offsets on the ground plane:
\begin{align}
    O^{t_0} = \mathrm{MLPHead}(\hat{Q}^{t_0}_{\mathrm{track}}) =
    \begin{bmatrix} \delta x, \delta y \end{bmatrix}^{T}.
\end{align}

\textbf{Heatmap Decoder.}
The Heatmap Decoder integrates the multi-scale Ground features of the current frame to predict human centers. 
It first employs a Feature Pyramid Network (FPN) to upsample and fuse $\hat{Q}^{t_0}$ into the highest spatial resolution. 
The fused representation is then passed through a convolutional regression head to predict the final crowd heatmap on the ground: $H^{t_0} = \mathrm{ConvHead}(\mathrm{FPN}(\hat{Q}^{t_0}))$.
% \begin{align}
%     H^{t_0} = \mathrm{ConvHead}(\mathrm{FPN}(\hat{Q}^{t_0})).
% \end{align}
Together, these two decoders jointly predict the crowd locations and their temporal displacements, enabling continuous multi-view tracking in dynamic environments.

%\zqnote{use mathematical language to describe the process of the step, how we do the view interaction. Try to follow the steps in transcenter and our module. Ignore the unrelated steps in Transcenter, just say we adopt [] as our encoder or decoder.}

\subsection{Model Training and Loss}
%\zqnote{If we have something to say; If not, it can be merged with the last step.}

The proposed model is trained by jointly optimizing a heatmap classification loss for both ground and image domains, and a regression loss for the motion offset on the ground plane. 
Following the uncertainty weighting strategy, we introduce two learnable parameters to adaptively balance the center and tracking branches during training.

\textbf{Heatmap Loss.}
To supervise the center prediction, we construct the ground-truth heatmap $H^{*}$ by placing Gaussian responses at each object center.
Given the predicted heatmap $H$, we apply the focal loss~\cite{lin2017focal}:
\begin{align}
    \mathcal{L}_{\mathrm{ground}} = \mathrm{FocalLoss}(H, H^{*}).
\end{align}
%$H^*$ means the ground-truth heatmap. 
An additional image-level supervision term $\mathcal{L}_{\mathrm{img}}$ is introduced using the same formulation to predict human center heatmaps in views.

\textbf{Offset Regression Loss.}
The offset decoder predicts the displacement of each tracked object center between consecutive frames.  
Given the predicted offset $O = [\delta x, \delta y]$ and its ground truth $O^{*}$, we employ an $\ell_1$ loss:
\begin{align}
    \mathcal{L}_{\mathrm{track}} = 
    \frac{1}{K}\sum_{x,y} 
    \|O_{xy} - O^{*}_{xy}\|_1,
    \quad \text{if } C^{*}_{xy} = 1.
\end{align}
This loss is only applied to valid center locations, ensuring sparse supervision over active tracks.

%\textbf{Uncertainty Weighting.}
\textbf{Total loss.}
To adaptively balance the losses of different branches, 
we follow the uncertainty weighting strategy~\cite{kendall2018multi}:
\begin{align}
    \mathcal{L}_{\mathrm{all}} 
    &\!=\! 10e^{-\sigma_c}\mathcal{L}_{\mathrm{ground}}\!
     +\! e^{-\sigma_t}\mathcal{L}_{\mathrm{track}}\!
     +\! \mathcal{L}_{\mathrm{img}}\!
     +\! \sigma_c \!+ \!\sigma_t,
\end{align}
where $\sigma_c$ and $\sigma_t$ are learnable uncertainty parameters 
for the center and tracking branches, respectively.  
This formulation allows the network to automatically calibrate the relative contribution of each branch during training.

\section{Experiments and Results}

\subsection{Datasets} %Collection and Generation

% \zqnote{Talk about how we collect the Canghai dataset and how we label it.
% Simply introduce how we label the CityStreet dataset and make it work for the multi-view tracking dataset}

% \textbf{Dataset Annotation.} 
% Our MVCrowdTrack and CityTrack datasets provide frame-level annotations, 
% including 2D object bounding boxes and ground-plane coordinates in a unified coordinate system (i.e., BEV annotations). 
% To facilitate efficient and accurate labeling, we developed a customized \textit{multi-camera annotation tool} 
% that allows annotators to visualize and adjust object positions across all camera views simultaneously. 
% We invited professional annotators to label both datasets using this tool. 
% After annotation, all data were carefully filtered and manually inspected 
% to ensure high accuracy and consistency of the annotations.

% \textbf{Calibration of the cameras}

\begin{figure}[t]
    \centering
    \includegraphics[width=0.9\linewidth]{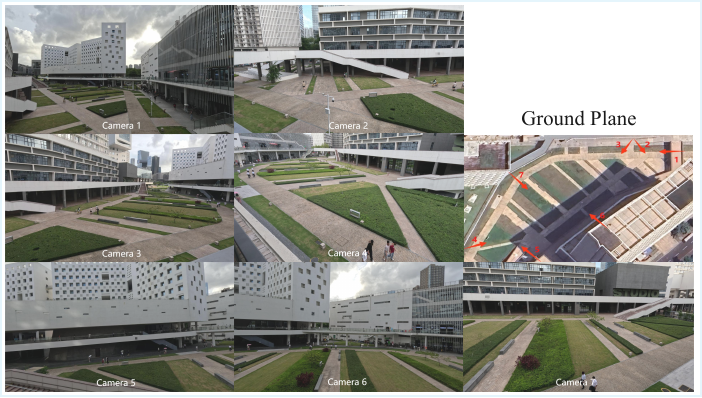}
    \vspace{-0.3cm}
    \caption{Multi-camera views and corresponding ground-plane layout in the MVCrowdTrack dataset.}
    \label{fig:dataset_vis}
    \vspace{-0.4cm}
\end{figure}

\textbf{MVCrowdTrack Dataset.} %MVCrowdTrack
To advance the multi-view crowd tracking task to more complicated conditions, we first collect and label a large real-world multi-view crowd tracking dataset,  MVCrowdTrack, which is collected on a large campus with a size of $120\,\text{m} \times 80\,\text{m}$. The scene of MVCrowdTrack is covered with 7 synchronized cameras together, with an image resolution of $5312\, \times 2988\,$, and lasts for 18 minutes. The frame rate of videos is 60, and we label 4 frames per second, resulting in a total of 4122 multi-view frames. In total, the dataset contains 342 people's trajectories with an average track length of 176 frames. In the experiments, 80\% of the data (3297 frames) are used for training, and the remaining 20\% (825 frames) are used for testing. We label the people in each view with bounding boxes, and make sure the same person is assigned a consistent ID across the time period. The foot points of each person are projected from all camera views onto the ground plane with camera calibrations (both extrinsic and intrinsic), and then the projected points from different views of the same person are averaged as the ground-truth location on the ground. The ground-plane map resolution is $1200 \times 800$, and 1 pixel is 0.1m in the real world. See an example in Figure \ref{fig:dataset_vis}.

\textbf{CityTrack Dataset.}
We have also labeled an existing multi-view crowd dataset called CityStreet~\cite{zhang2019wide} for the multi-view tracking task. 
% CityStreet consists of 500 frames captured from 3 camera views, with a time gap of 2 seconds between neighboring frames, covering a scene size of $64\,\text{m} \times 76\,\text{m}$. It only provides ground plane locations as ground truth labels, without consistent IDs between neighbour frames for the same person.
CityTrack contains 2,588 frames from the CityStreet dataset videos, annotated at 4 fps, and ensures that the IDs of the crowds are consistent for tracking tasks. It contains trajectories with an average track length of 228 frames. The first 1948 frames are used for training, and the rest are used for testing. The ground-plane resolution is $640 \times 768$, and 1 pixel represents $0.1\,\text{m}$ in the real world.

Compared to \textbf{Wildtrack} and \textbf{MultiviewX} in Table~\ref{tab:dataset_comparison}, the newly proposed \textbf{MVCrowdTrack} and \textbf{CityTrack} datasets have larger scene scales, higher spatial resolutions, denser frame annotations, and significantly longer average trajectory lengths. Specifically, MVCrowdTrack covers a wider area with higher-resolution cameras for a longer period, while CityTrack also provides denser temporal annotations and many more crowds with consistent IDs. These characteristics make them more suitable for evaluating multi-view crowd tracking approaches in real-world and complex scenarios, where existing small-scale datasets may not fully reflect the challenges of practical applications.

% denoted as the datasetlong samplessamples
%  dataset by applying  calibration parameters. To obtain the final ground-truth location,corresponding tocorresponding to a physical area of $120\,\text{m} \times 80\,\text{m}$

% ed}2,588 at4\,fps more labeledpedestriansactual  The first 1948 frames are used for training, and the remaining frames are used for testing. campus of $120\,\text{m} \times 80\,\text{m}$ seven synchronizedconsistent pedestrian identities on the CityStreet videos

\begin{table}[t]
\centering
\scriptsize
\caption{Comparison of multi-view pedestrian tracking datasets.}
%\resizebox{
% \columnwidth}{!}{
 \vspace{-0.2cm}
\begin{tabular}{@{\hspace{0cm}}l@{\hspace{0.1cm}}c@{\hspace{0.1cm}}c@{\hspace{0.1cm}}c@{\hspace{0.1cm}}c@{\hspace{0.1cm}}c@{\hspace{0.1cm}}c@{\hspace{0.1cm}}c@{\hspace{0.1cm}}}
\hline
Dataset & Resolution & View & People & Frame & FPS & Size ($m^2$) & Avg. Track Len. \\
\hline
MultiviewX   & 1920×1080   & 6 & 360 & 400  & 2 & 25×16 & 44 \\
Wildtrack    & 1920×1080   & \textbf{7} & 313 & 400  & 2 & 36×12 & 30 \\
CityTrack    & 2704×1520   & 3 & \textbf{950} & 2588 & \textbf{4} & 64×76 & \textbf{228} \\
MVCrowdTrack & \textbf{5312×2988} & \textbf{7} & 342 & \textbf{4122} & \textbf{4} & \textbf{120×80} & 176 \\
\hline
\end{tabular}
%}
 \vspace{-0.6cm}
\label{tab:dataset_comparison}
\end{table}

\begin{table*}[t]
\small
\centering
\caption{Comparison of the multi-view crowd tracking performance on the larger datasets MVCrowdTrack and CityTrack using 5 metrics. 
%The distance threshold is $1$m on MVCrowdTrack (5 pixels on the ground plane map), and $2$m on CityStreet (20 pixels on the ground plane map).  See results with other distance thresholds in the supplemental.
%Overall, all previous methods do not perform well on the 2 large datasets compared to Wildtrack and MultiviewX (see in Table \ref{table:Wildtrack_results}).
The proposed method ranks the best among all methods according to the metrics on the two datasets. We use MOTA and IDF1 as main metrics.
\textbf{Bold} font indicates the best metric, and \underline{underline} font indicates the second-best.
}
 \vspace{-0.3cm}
%\begin{tabular}{l@{\hspace{0.12cm}}|c@{\hspace{0.12cm}}c@{\hspace{0.12cm}}c@{\hspace{0.12cm}}c@{\hspace{0.12cm}}c@{\hspace{0.12cm}}|c@{\hspace{0.12cm}}c@{\hspace{0.12cm}}c@{\hspace{0.12cm}}c@{\hspace{0.12cm}}c@{\hspace{0.12cm}}}
\begin{tabular}{l|ccccc|ccccc}%{l@{\hspace{0.15cm}}|c@{\hspace{0.15cm}}c@{\hspace{0.15cm}}c@{\hspace{0.15cm}}c@{\hspace{0.15cm}}c@{\hspace{0.15cm}}|c@{\hspace{0.15cm}}c@{\hspace{0.15cm}}c@{\hspace{0.15cm}}c@{\hspace{0.15cm}}c@{\hspace{0.15cm}}}
\hline
    Dataset &  \multicolumn{5}{c|}{MVCrowdTrack}  &  \multicolumn{5}{c}{CityTrack}  \\
%\hline
    Method & MOTA$\uparrow$ & MOTP$\uparrow$ & IDF1$\uparrow$ & MT$\uparrow$ & ML$\downarrow$  & MOTA$\uparrow$ & MOTP$\uparrow$ & IDF1$\uparrow$ & MT$\uparrow$ & ML$\downarrow$ \\
\hline
    EarlyBird \cite{Teepe_2024_WACV}    & 54.56    & 30.46   & 53.84
  & 24.48  & 14.22  
                 & \underline{48.85}  & 21.83    & 32.15     & 17.33     & \underline{13.9}     \\  
    MVFlow \cite{engilberge2023multi}      & 49.82  & \textbf{46.79}     & 44.06    & 22.22      & 37.04 
                 & 38.19  & 6.94    & 27.89     & 8.92     & 24.88   \\ 
    TrackTacular \cite{Teepe_2024_CVPR} & \underline{62.86}    & 29.23     & \underline{58.71}     & \underline{40.81}  & \underline{10.20}
  
                 & 43.37  & \textbf{23.23}  & \underline{32.49} & \underline{20.43}     & \textbf{12.38}      \\
    %ReST \cite{cheng2023rest}        & -    & -     & -    & -      & -                  & -     & -    & -   & -    & -    \\ 
\hline
\hline
     MVTrackTrans (Ours)        & \textbf{63.87}    & \underline{40.59}     & \textbf{59.06}    & \textbf{42.85}      & \textbf{8.16}  
                 & \textbf{55.39} & \underline{22.71} & \textbf{34.41} & \textbf{25.07} & \underline{12.69}   \\
\hline
\end{tabular}
\vspace{-0.4cm}
%\vspace{-0.4cm}
\label{table:table_results}
\end{table*}

\begin{table}[t]
\centering
\small
%\scriptsize 
\caption{Tracking performance on CityTrack for different variants of our proposed MVTrackTrans model.}
\vspace{-0.3cm}
\begin{tabular}{@{\hspace{0cm}}l@{\hspace{0.08cm}}c@{\hspace{0.12cm}}c@{\hspace{0.12cm}}c@{\hspace{0.12cm}}c@{\hspace{0.12cm}}c@{\hspace{0cm}}}
\hline
Method & MOTA$\uparrow$ & MOTP$\uparrow$ & IDF1$\uparrow$ & MT$\uparrow$ & ML$\downarrow$ \\
\hline
Baseline & 54.92 & \textbf{22.83} & 34.11 & 27.86 & 13.00 \\
 + View Prediction Branch & 53.17 & 20.68 & 32.65 & \textbf{30.34} & \textbf{10.52} \\
 ++ ViewInteraction (Ours) & \textbf{55.39} & 22.71 & \textbf{34.41 }& 25.07 & 12.69 \\
\hline
\end{tabular}
\vspace{-0.5cm}
\label{tab:trackformer_variants}
\end{table}

\subsection{Experiment Settings}

\textbf{Comparison methods.} 
We have compared with the state-of-the-art multi-view crowd tracking methods, such as 
Earlybird \cite{Teepe_2024_WACV}, MVFlow \cite{engilberge2023multi}, and TrackTacular \cite{Teepe_2024_CVPR} on the proposed MVCrowdTrack and CityTrack datasets. The code provided for each method is adopted from their papers, and they are trained and evaluated with the same settings as our model. We have tried to run ReST \cite{cheng2023rest} on the new datasets, but it failed. 
For other methods like MVTrajecter \cite{Yamane_2025_ICCV}, MVTr \cite{yang2024end} or MCBLT \cite{Wang_2025_ICCV}, since no codes are provided, we cannot compare with them on the two new datasets.
%and the existing two small datasets, MultiviewX \cite{hou2020multiview} and Wildtrack \cite{chavdarova2018wildtrack}.
We also compare our model with other existing methods, REMP \cite{11018739}, MCBLT \cite{Wang_2025_ICCV}, MVTr \cite{yang2024end} on MultiviewX and Wildtrack. Their metrics on MultiviewX and Wildtrack are adopted from their papers.
%\zqnote{udpate later}

\textbf{Implementation details.} 
Our framework follows the ResNet18 \cite{he2016deep} feature extractor and the Transformer encoder–decoder architecture presented in Deformable DETR \cite{zhu2020deformable}. During training, all input images are resized to 1280 × 720 pixels. We train for 50 epochs on all datasets, including MVCrowdTrack and CityTrack. The initial learning rate is 0.01. All experiments are conducted on 4 NVIDIA RTX 4090 GPUs with a batch size of 1.

\textbf{Evaluation metrics.} 
All tracking metrics are evaluated on the ground plane to ensure consistent spatial alignment. We adopt the same standard Multiple Object Tracking (MOT) metrics as the latest SOTAs \cite{Teepe_2024_CVPR, Teepe_2024_WACV} along with identity-aware measures. The distance threshold for positive association is set to $r = 2\,\text{m}$ for the larger-scale MVCrowdTrack and CityTrack datasets, and  $r = 1\,\text{m}$ on the Wildtrack and MultiviewX datasets. \textit{The main evaluation metrics are Multiple Object Tracking Accuracy (MOTA) and IDF1}, which jointly consider missed detections, false positives, and identity switches. We further report Mostly Tracked (MT) and Mostly Lost (ML), representing the proportion of trajectories that are successfully tracked for more than 80\% or less than 20\% of their lifespan, respectively, relative to the total number of unique pedestrians in the test set.

\subsection{Experiment Results}
%\textbf{MVCrowdTrack.} 
%\zqnote{Talk about experiments on Canghai and analyze the results: better or not, why, according to the table and figures.}
We show the multi-view tracking performance on MVCrowdTrack and CityTrack in Table \ref{table:table_results}.
On \textbf{MVCrowdTrack}, our proposed method MVTrackTrans achieves the best performance across all methods. 
%in terms of MOTA, IDF1, MT, and ML metrics, and the second best MOTP.
EarlyBird is a typical CNNs-based multi-view crowd tracking method supervised with similar heatmaps to ours. But it is much worse compared to our method on MVCrowdTrack, demonstrating the transformer architecture's advantages on large and complicated scenes.
MVFlow achieves the worst performance, where the possible reason is that it uses a weakly-supervised human motions for tracking, resulting in much lower metrics for long-time tracking on MVCrowdTrack.
% the flow estimation error of the method would be accumulated for a longer time tracking task, resulting in lower metrics on MVCrowdTrack.
TrackTacular uses better historical information compared to EarlyBird. It achieves the second-best IDF1, MOTA, MT, and ML metrics, but it is still worse than our method. Generally, our method is the best among all methods, proving that the transformer model's superiority on the large and long-time multi-view crowd tracking tasks.

On \textbf{CityTrack}, the proposed method also achieves the best performance according to MOTA, IDF1, and MT metrics, and the second best MOTP and ML.
Compared to MultiviewX and Wildtrack, CityTrack is a dataset with larger sizes, more crowds, and more severe occlusions. Our method achieves the best results, revealing that our method can handle these challenges better than existing methods.
EarlyBird is the second according to MOTA, showing that it is quite a stable architecture across different datasets, but it is behind our transformer model, due to which our model can be better adapted to more complicated scenarios.
Similarly, MVFlow performs the worst on CityTrack for similar reasons as on MVCrowdTrack.
TrackTacular achieves the second IDF1, but much lower MOTA than ours, which suggests that TrackTacular cannot perform detection well on complicated datasets.

\begin{figure*}[t]
    \centering
    \includegraphics[width=0.88\linewidth]{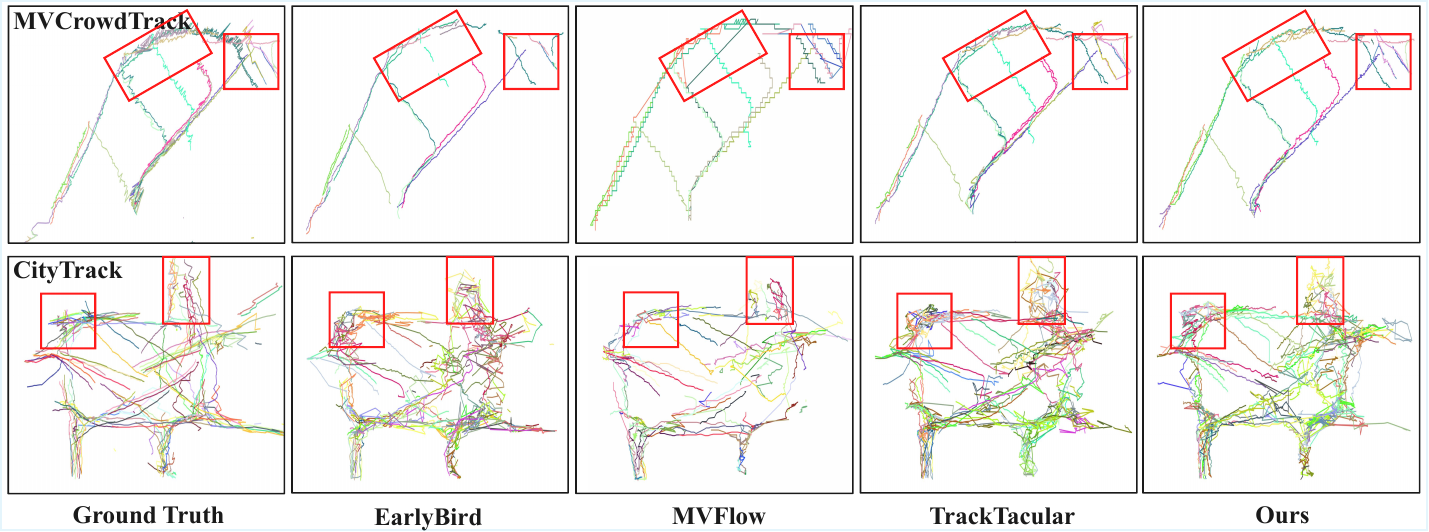}
    \vspace{-0.3cm}
    \caption{The predicted trajectory visualizations. Our method can accurately track more people for a long time (see red boxes).}
    \label{fig:result_vis}
    \vspace{-0.6cm}
\end{figure*}

Overall, the proposed MVTrackTrans method achieves the best performance among all methods on the two large real-world datasets.
The reason is that we adopt a transformer-based architecture for the large and complicated scenes, which provides stronger spatial (multi-view) and temporal fusion for the tracking task. In addition, the proposed view-ground interaction module further improves the tracking results (see the ablation study in Sec. 4.4).
We also show the \textbf{visualization} results of predicted trajectories on the MVCrowdTrack and CityTrack datasets in Figure~\ref{fig:result_vis}. It concludes that our method can accurately track more people compared to other methods as seen in the red boxes. Especially, as shown in the red box of the results on MVCrowdTrack, after a long time period of tracking, our method could still track more people compared to EarlyBird and TrackTacular. %Note that MVFlow adopts a ground-plane transformation whose output resolution is inconsistent with others, therefore, we do not visualize it here.%

%\zqnote{Add a figure showing results on the 2 datasets.}
%\zqnote{Xinquan.}

\subsection{Ablation Study}

\textbf{Architecture ablation study.} 
We have conducted an ablation study on the model architecture on the CityTrack dataset: `Baseline', `+View Prediction Branch', `++ViewInteractions (Ours)', in Table \ref{tab:trackformer_variants}.
`Baseline' means no 2D camera view heatmap prediction branch or supervision is used in the model training; `+View Prediction Branch' means adding the 2D heatmap prediction branch to the Baseline model, but without the proposed view-ground interaction module; `++ViewInteractions (Ours)' means further adding our proposed view-ground interaction module to the model.
%\zqnote{explain each model first.}
%Pure Trackformer baseline, Trackformer+2D branch (2D loss training), Trackformer+2D branch+ViewGroundInteract
From Table \ref{tab:trackformer_variants}, we conclude that simply adding the 2D camera-view branch in the model does not improve the model performance due to the competition in the 2D and ground-plane task training. And with our view-ground interaction module further, the model can achieve better tracking performance, especially according to MOTA and IDF1 metrics. The reason is that the proposed view-ground interaction module fuses both the camera view and ground information, which helps to achieve more stable tracking for a long time.

\begin{table}[t]
\centering
\small
%\scriptsize 
\caption{Ablation study on the view-ground interaction module, which is conducted on the CityTrack dataset.
% \zqnote{list different ways in the view-ground interaction}
}
\vspace{-0.3cm}
\begin{tabular}{@{\hspace{0cm}}l@{\hspace{0.1cm}}ccccc@{\hspace{0cm}}} 
\hline
Method & MOTA$\uparrow$ & MOTP$\uparrow$ & IDF1$\uparrow$ & MT$\uparrow$ & ML$\downarrow$ \\
\hline
SelfAtt & 55.38 & \textbf{23.71} & 33.64 & \textbf{27.86} & \textbf{12.38} \\
CrossAtt (Ours) & \textbf{55.39} & 22.71 & \textbf{34.41} & 25.07 & 12.69 \\
\hline
\end{tabular}
\label{tab:view-ground module}
\vspace{-0.3cm}
\end{table}

\begin{table}[t]
\centering
\small
%\scriptsize 
\caption{Ablation study on the supervision manner:
Use sparse queries with direct coordinate regression loss, or use dense pixel representations with heatmap prediction loss (ours). The experiment is conducted on the CityTrack dataset
% \zqnote{use trackformer ((Keep the MV tracking model based on transcenter, and replace the loss with this)) and transcenter loss for comparison, proving that the pixel-wise loss is better for the multi-view crowd tracking task (probably due to the multi-view fusion step, and density map loss is stronger and more effective)}
}
\vspace{-0.3cm}
\begin{tabular}{@{\hspace{0cm}}l@{\hspace{0.05cm}}c@{\hspace{0.12cm}}c@{\hspace{0.12cm}}c@{\hspace{0.12cm}}c@{\hspace{0.12cm}}c@{\hspace{0cm}}}
\hline
Training & MOTA$\uparrow$ & MOTP$\uparrow$ & IDF1$\uparrow$ & MT$\uparrow$ & ML$\downarrow$ \\
\hline
Coordinate regression   & 40.71 & 13.88 & 31.45 & 9.28 & 20.12 \\ %\cite{meinhardt2022trackformer}
Heatmap regression  (Ours) & \textbf{55.39} & \textbf{22.71} & \textbf{34.41} & \textbf{25.07} & \textbf{12.69} \\
\hline %\cite{9964258}
\end{tabular}
\label{tab:training}
\vspace{-0.5cm}
\end{table}

\textbf{View-ground interaction ablation study.} 
%How we perform the view-ground interactions: simple method, xxx, our method.
We conduct an ablation study on the view-ground interaction module in Table \ref{tab:view-ground module}.
We compare different ways of implementing the module: `SelfAtt', and `CrossAtt (Ours)'.
`SelfAtt' means the camera view queries are fused with the ground queries with a self attention mechanism; `CrossAtt (Ours)'  the camera view queries are fused with the ground queries with a cross attention mechanism. 
As in Table \ref{tab:view-ground module}, the best performance is achieved by using cross attention, in terms of MOTA and IDF1.
The reason is that the cross attention provides a more thorough fusion of the camera view features and ground features, compared to self attention, resulting in enhanced performance.

\textbf{Training method ablation study.} 
%How we perform the view-ground interactions: simple method, xxx, our method.
We also conduct an ablation study on the model training in Table \ref{tab:training}.
We compare different ways of training our proposed transformer-based multi-view crowd tracking model: coordinate regression and heatmap regression.
The first use spare queries and direct coordinate ground-truth as supervision, as in 2D transformer tracking method \cite{meinhardt2022trackformer};
The second one uses dense pixel representations with heatmap prediction loss (Ours) for supervision, as in \cite{9964258}.
As in Table \ref{tab:training}, the model trained with heatmap regression is much better than using the direct coordinate regression. The reason might be that in the multi-view crowd tracking task, the multi-view projection step stretches the features on the ground, which causes extra difficulties for accurate tracking. And the dense heatmap supervision can better guide the model training to reject these noises, and thus better performance is achieved.

\begin{table}[t]
\centering
\scriptsize
\caption{Comparison with previous methods on Wildtrack. For MCBLT, we report the results using its original detector.}
\vspace{-0.2cm}
\begin{tabular}{l@{\hspace{0.15cm}}c@{\hspace{0.15cm}}c@{\hspace{0.15cm}}c@{\hspace{0.15cm}}c@{\hspace{0.15cm}}c@{\hspace{0.15cm}}}
\hline
{Method} & {MOTA$\uparrow$} & {MOTP$\uparrow$} & {IDF1$\uparrow$} & {MT$\uparrow$} & {ML$\downarrow$} \\
\hline
KSP-DO \cite{chavdarova2018wildtrack} & 69.6 & 61.5 & 73.2 & 28.7 & 25.1 \\
KSP-DO-ptrack \cite{chavdarova2018wildtrack} & 72.2 & 60.3 & 78.4 & 42.1 & 14.6 \\
GLMB-YOLOv3 \cite{ong2020bayesian} & 69.7 & 73.2 & 74.3 & 79.5 & 21.6 \\
GLMB-DO \cite{ong2020bayesian} & 70.1 & 63.1 & 72.5 & \underline{93.6} & 22.8 \\
DMCT \cite{you2020real} & 72.8 & 79.1 & 77.8 & 61.0 & \underline{4.9} \\
DMCT Stack \cite{you2020real} & 74.6 & 78.9 & 81.9 & 65.9 & \underline{4.9} \\
ReST\textsuperscript{\dag}~\cite{cheng2023rest} & 81.6 & 81.8 & 85.7 & 79.4 & \underline{4.9} \\
MCBLT\cite{Wang_2025_ICCV} & 87.5 & \textbf{94.3} & 93.4 & 90.2 & \textbf{2.4} \\
REMP\textsuperscript{\dag}~\cite{11018739} & 88.5 & 86.8 & -- & -- & -- \\
EarlyBird~\cite{Teepe_2024_WACV} & 89.5 & 86.6 & 92.3 & 78.0 & \underline{4.9} \\
MVFlow~\cite{engilberge2023multi} & 91.3 & -- & 93.5 & -- & -- \\
TrackTacular~\cite{Teepe_2024_CVPR} & 91.8 & 85.4 & 95.3 & 87.8 & \underline{4.9} \\
MVTr~\cite{yang2024end} & 92.3 & 92.7 & 93.1 & \textbf{95.1} & \underline{4.9} \\
MVTrajecter \cite{Yamane_2025_ICCV} & \textbf{94.3} & \underline{93.0} & \underline{96.5} & 90.2 & \underline{4.9} \\
%MCBLT \textsuperscript{\dag}~\cite{Wang_2025_ICCV} & 92.6 & 93.7 & 95.6 & 80.5 & 7.3 \\
\hline
% MVTrackTrans (Ours) & 91.17 & 86.87 & 94.11 & 82.92 & 4.87 \\
MVTrackTrans (Ours) & 91.2 & 86.9 & 94.1 & 82.9 & \underline{4.9} \\
MVTrackTrans (Ours)$^{\dagger}$ & \underline{93.6} & 86.7 & \textbf{96.7} & 85.4 & \underline{4.9} \\
\hline
\end{tabular}
\vspace{-0.6cm}
\label{tab:wildtrack_results}
\end{table}

\textbf{Performance on small datasets Wildtrack and MultiviewX.} 
We have also conducted experiments on small datasets, Wildtrack and MultiviewX, as shown in Table \ref{tab:wildtrack_results} and \ref{tab:multiviewx_results}, respectively. The proposed model outperforms several methods on Wildtrack, such as ReST, MCBLT, EarlyBird, and REMP. While it achieves comparable performance on Wildtrack as SOTAs TrackTacular, MVTr, or MVFlow, though lower than MVTrajecter. It also achieves comparable performance on MultiviewX as EarlyBird, but lower performance than MVTrajecter, TrackTacular and MVTr. 
{\color{black}Additionally, MVTrackTrans$^{\dagger}$ denotes a variant of MVTrackTrans with two-frame temporal feature fusion and Kalman filter to improve temporal consistency, achieving performance comparable to existing SOTAs.} Overall, MVTrackTrans also achieves good results on small datasets. 

\begin{table}[t]
\centering
\scriptsize
\caption{Comparison with previous methods on MultiviewX. 
%The method with $\dagger$ used external data or external models.
}
\vspace{-0.2cm}
\begin{tabular}{l@{\hspace{0.15cm}}c@{\hspace{0.15cm}}c@{\hspace{0.15cm}}c@{\hspace{0.15cm}}c@{\hspace{0.15cm}}c@{\hspace{0.15cm}}}
\hline
{Method} & {MOTA$\uparrow$} & {MOTP$\uparrow$} & {IDF1$\uparrow$} & {MT$\uparrow$} & {ML$\downarrow$} \\
\hline
REMP$^{\dagger}$~\cite{11018739} & 81.0 & 85.8 & -- & -- & -- \\
EarlyBird~\cite{Teepe_2024_WACV} & 88.4 & 86.2 & 82.4 & 82.9 & \underline{1.3} \\
TrackTacular~\cite{Teepe_2024_CVPR} & \underline{92.4} & 80.1 & 85.6 & 92.1 & 2.6 \\
MVTr~\cite{yang2024end} & 91.4 & \textbf{95.0} & 82.9 & \underline{96.1} & \textbf{0.0} \\
MVTrajecter \cite{Yamane_2025_ICCV} & \textbf{92.8} & \textbf{95.0} & \underline{85.8} & \textbf{97.4} & \textbf{0.0} \\
\hline
MVTrackTrans (Ours) & 89.8 & \underline{90.2} & 72.1 & 85.5 & 6.6 \\
MVTrackTrans (Ours)$^{\dagger}$ & 90.2 & 83.6 & \textbf{86.3} & 88.2 & 3.9 \\
\hline
\end{tabular}
\vspace{-0.6cm}
\label{tab:multiviewx_results}
\end{table}

\section{Conclusion}

In this paper, we aim to advance the current study for the multi-view crowd tracking task to more challenging and practical scenarios. Thus, first, we propose to collect a large multi-view tracking dataset that contains a much larger scene size with a long time period, and label an existing large real-world multi-view crowd dataset for the task.
Besides, instead of exploring pure CNN-based models as previous research in the area, we propose a transformer-based multi-view crowd tracking model, \textit{MVTrackTrans}, which adopts interactions between camera views and the ground plane for better multi-view tracking. Compared with existing methods on the two large real-world datasets, the proposed MVTrackTrans model achieves much better performance. We believe the proposed datasets and model will advance the multi-camera-based multi-object tracking task to more practical scenarios.

\section*{Acknowledgments}
\small
This work was supported in part by Guangdong Science and Technology Program (2024B0101050004), NSFC (62202312), ICFCRT (W2441020), Shenzhen Science and Technology Program (KJZD20240903100022028, KQTD20210811090044003), Scientific Foundation for Youth Scholars and Scientific Development Funds from Shenzhen University.

%\zqnote{update later}

% Multi-view crowd tracking estimates each person's tracking trajectories on the ground of the scene. Recent research works rely on designing CNNs-based multi-view crowd tracking architectures, and most of them are evaluated and compared on relatively small datasets, such as Wildtrack and MultiviewX. Since these two datasets are collected in small scenes and only contain tens of frames in the evaluation stage, it is difficult for the current methods to be applied to real-world applications where scene size and occlusion are more complicated. In this paper, we propose a Transformer-based multi-view crowd tracking model, \textit{MVTrackFormer}, which adopts interactions between camera views and the ground plane for better multi-view tracking. To better evaluate the methods, we propose to collect a large multi-view tracking dataset that contains a much larger scene size with a long time period, and label an existing multi-view crowd dataset for the task. Compared with existing methods on the two large and new datasets, the proposed MVTrackFormer model achieves much better performance. We believe the proposed datasets and model will push the frontiers of the area to more practical scenarios, and the dataset and code will be made public upon paper acceptance.

{
    \small
    \bibliographystyle{ieeenat_fullname}
    \bibliography{main}
}

% WARNING: do not forget to delete the supplementary pages from your submission 
% \input{sec/X_suppl}

\end{document}